\documentclass[sigconf]{aamas-arxiv}

\usepackage{balance} % for balancing columns on the final page
\usepackage[procnumbered,linesnumbered,ruled,vlined]{algorithm2e}
\usepackage{amsmath}

\usepackage{subcaption}

\usepackage{cleveref}

\title{Improving Causal Effect Estimation of Weighted Regression Based Estimator using Neural Networks}

\author{Plabon Shaha}
\authornote{Both authors contributed equally to the paper}
\affiliation{
  \institution{University of Dhaka}
  \city{Dhaka}
  \country{Bangladesh}}

\author{Talha Islam Zadid}
\authornotemark[1]
\affiliation{
  \institution{University of Dhaka}
  \city{Dhaka}
  \country{Bangladesh}}

\author{Ismat Rahman}
\affiliation{
  \institution{University of Dhaka}
  \city{Dhaka}
  \country{Bangladesh}}
  
\author{Md. Mosaddek Khan}
\affiliation{
  \institution{University of Dhaka}
  \city{Dhaka}
  \country{Bangladesh}}

\begin{abstract}
Estimating causal effects from observational data supports decisions about interventions in healthcare, industry, and public policy. Even when a causal effect is identifiable, estimating its associated statistical functional from finite samples can be challenging, particularly with high-dimensional covariates. Composition of weighting operators (CWO) provides a framework for estimating certain effects beyond ordinary back-door adjustment. We investigate a neural-network regression implementation of this framework, NN-CWO, to approximate the regression functions appearing in these estimands. We compare NN-CWO with the parametric CWO baseline in front-door, surrogate, and multi-outcome sequential back-door simulation settings. The reported experiments show lower median average absolute error for NN-CWO in the evaluated settings, with the original evaluation reporting improvements of up to approximately $55\%$.
\end{abstract}


\newcommand{\BibTeX}{\rm B\kern-.05em{\sc i\kern-.025em b}\kern-.08em\TeX}

\begin{document}
\makeatletter
\let\@authorsaddresses\@empty
\makeatother

\settopmatter{printacmref=false} % Removes citation information below abstract
\renewcommand\footnotetextcopyrightpermission[1]{} % removes footnote with conference information in first column
\pagestyle{plain} % removes running headers

%%% The following commands remove the headers in your paper. For final 
%%% papers, these will be inserted during the pagination process.

\pagestyle{fancy}
\fancyhead{}

%%% The next command prints the information defined in the preamble.

\maketitle 

%%%%%%%%%%%%%%%%%%%%%%%%%%%%%%%%%%%%%%%%%%%%%%%%%%%%%%%%%%%%

\section{Introduction}

Causal inference studies how interventions change outcomes. Structural causal models encode assumptions about the data-generating process and distinguish observational associations from intervention effects~\cite{Pearl_1995,pearl-2009}. Their causal interpretation depends on the validity of these assumptions; using a causal model does not by itself guarantee robustness to changes in the environment. Applications include medicine~\cite{medi-2,medi-5}, Mendelian randomization~\cite{gene-6}, public health and epidemiology~\cite{business-8,epidem-9}, economics~\cite{business-9,econ-12,econ-13}, and neighborhood health~\cite{social-11}.

Identification and estimation are distinct tasks. Identification asks whether a causal target is uniquely determined by the available population distribution and causal assumptions. The do-calculus provides rules for deriving identification formulas from a causal graph~\cite{Pearl_1995}. Estimation then approximates an identified functional using finite data. Limited samples can make this approximation inaccurate without invalidating an identification result. The back-door (BD) criterion is a sufficient graphical condition for identification by covariate adjustment, rather than a remedy specific to small samples~\cite{pearl-2009}.

Directly estimating a large collection of conditional probabilities can be difficult when covariates are high-dimensional. Weighting and regression provide alternatives to such enumeration. Inverse-probability weighting has roots in unequal-probability survey sampling~\cite{horvitz1952generalization}; its causal use additionally requires appropriate identifying and support conditions. For binary treatment, the propensity score is the conditional treatment probability given observed covariates~\cite{rosenbaum1983central}. Adjustment for this score can address measured confounding under the relevant assumptions, but does not generally remove unmeasured confounding. Matching methods also seek balance in observed covariates~\cite{gene-7}.

Inverse-probability weighting for marginal structural models can address measured time-varying confounders affected by earlier treatment, subject to the requisite assumptions~\cite{hernan2000marginal}. Stabilization changes the form of the weights; it does not remove those assumptions. Doubly robust estimators combine treatment and outcome models and can remain consistent if either model is correctly specified, with the other identifying and regularity conditions in place~\cite{bang2005doubly}. Flexible outcome regression, including Bayesian nonparametric modeling, is another approach to causal estimation under suitable identification assumptions~\cite{hill2011bayesian}. The g-formula supports identification and estimation of effects of treatment sequences when its conditions hold~\cite{robins1986new}.

Some effects are identifiable even when ordinary BD adjustment for the target treatment and outcome is unavailable. The usefulness of mediators, surrogate variables, or sequential covariates depends on their causal positions and the corresponding graphical conditions. Jung et al.~\cite{jung2020estimating} express several such estimands using weighting operators and their composition, including front-door, surrogate, and multi-outcome sequential back-door settings. Related work connects causal estimation with weighted empirical risk minimization~\cite{jung2020learning}.

We retain the CWO identification formulas and study a neural-network implementation of their regression components. The comparison concerns the parametric regression specifications used by the CWO baseline, rather than regression in general: neural-network fitting with squared loss is itself regression. Nonlinearities and interactions may challenge a restricted model class, but increasing dimension does not necessarily make a relationship nonlinear. NN-CWO uses backpropagation to fit regression functions within an identified causal calculation; it does not discover causal structure. Our contribution is this empirical implementation and its evaluation in three simulated settings.

Section~\ref{sec:bg} defines the causal targets and weighting identities. Section~\ref{sec:methods} describes NN-CWO, Section~\ref{sec:res} presents the comparisons, and Section~\ref{sec:conc} discusses their scope and future work.

%%%%%%%%%%%%%%%%%%%%%%%%%%%%%%%%%%%%%%%%%%%%%%%%%%%%%%%%%%%%%%%%%%%%%%%%

\section{Background and Problem definition}
\label{sec:bg}
We consider acyclic structural causal models $\mathcal M=(U,V,F,P_U)$, with equations $V_i=f_i(Pa_i,U_i)$, where $Pa_i\subseteq V\setminus\{V_i\}$ and $U_i\subseteq U$. Here $V$ denotes the observed endogenous variables in the models considered, $U$ denotes exogenous variables, and $P_U$ their joint distribution. Endogenous and observed are not synonymous in general. Directed edges encode structural dependence; bidirected edges in the observed-variable graph represent latent common causes.

For a given query, partition the observed variables into treatment, covariate, and outcome blocks, $V=\mathbf X\mathbin{\dot\cup}\mathbf Z\mathbin{\dot\cup}\mathbf Y$. The intervention $do(\mathbf X=\mathbf x)$ replaces the treatment equations by constants. Write $P_{\mathbf x}(\mathbf y)=P(\mathbf Y=\mathbf y\mid do(\mathbf X=\mathbf x))$. This distribution is identifiable if all models compatible with the assumed graph and observational distribution agree on it~\cite{pearl-2009}. The sample size concerns estimation of this target, not its definition or identification.

Uppercase letters denote variables, lowercase letters their values, and bold letters variable blocks. We suppress boldface where the distinction is clear. For scalar outcomes, define
\[
\mu(x)=\mathbb E_{P_x}[Y],\qquad
\mu_h(x)=\mathbb E_{P_x}[h(\mathbf Y)].
\]
Hats denote sample-based estimates. Sums below denote discrete marginalization and are replaced by integrals for continuous variables. We use $Pa$, $Ch$, $An$, and $De$ for parents, children, ancestors, and descendants. In $G_{\overline A\underline B}$, incoming edges into $A$ and outgoing edges from $B$ are removed.

\theoremstyle{definition}
\begin{definition}[Weighted distribution]
For a positive weight function $\omega(v)$ satisfying $0<\mathbb E_P[\omega(V)]<\infty$, define
\[
p^\omega(v)=\frac{p(v)\omega(v)}{\mathbb E_P[\omega(V)]}.
\]
\label{def_weight_distribution}
\end{definition}
The symbol $\omega$ distinguishes a weight from the surrogate variable $W$ used below.

\begin{definition}[Back-door adjustment]
If $\mathbf Z$ satisfies the BD criterion relative to $(\mathbf X,Y)$, then, on the required support,
\begin{equation}
\label{back_eq}
\begin{aligned}
p_{\mathbf x}(y)&=\sum_{\mathbf z}p(y\mid\mathbf x,\mathbf z)p(\mathbf z),\\
\mu(\mathbf x)&=\sum_{\mathbf z}\mathbb E[Y\mid\mathbf X=\mathbf x,\mathbf Z=\mathbf z]p(\mathbf z).
\end{aligned}
\end{equation}
\end{definition}
The classical BD criterion excludes descendants of the treatment from the adjustment set and requires that it block paths entering the treatment through an incoming arrow~\cite{pearl-2009}.

\subsection{Weighting Operator and the Formulation of Non Back-door Scenarios}
\begin{definition}[Weighting operator]
For regression inputs $A$ and response function $h(Y)$, define
\begin{equation}
\label{weight_operator}
\begin{aligned}
\beta[h(Y)\mid a;\omega]
&=\mathbb E_{P^\omega}[h(Y)\mid A=a]\\
&=\frac{\mathbb E_P[\omega(V)h(Y)\mid A=a]}
{\mathbb E_P[\omega(V)\mid A=a]}.
\end{aligned}
\end{equation}
\end{definition}
This is a statistical functional. It equals a causal target only for appropriate weights and identifying conditions. For example, under BD adjustment,
\[
\omega(v)=\frac{p(x)}{p(x\mid z)},\qquad
\mu(x)=\beta[Y\mid x;\omega].
\]
The weighted conditional outcome mean then equals the interventional mean; the entire weighted joint distribution need not equal the joint distribution under a fixed intervention.

The following identities require the stated graphical conditions and positivity of the probabilities used in their denominators on the relevant support. Estimation additionally depends on the probability and regression models. Very small estimated probabilities can yield unstable weights, and flexible outcome regression alone does not remedy this problem or guarantee correctness under misspecified weight models. We use the CWO operators~\cite{jung2020estimating} for the three settings below.

\begin{figure}[t]
    \centering
    \includegraphics[width=\linewidth]{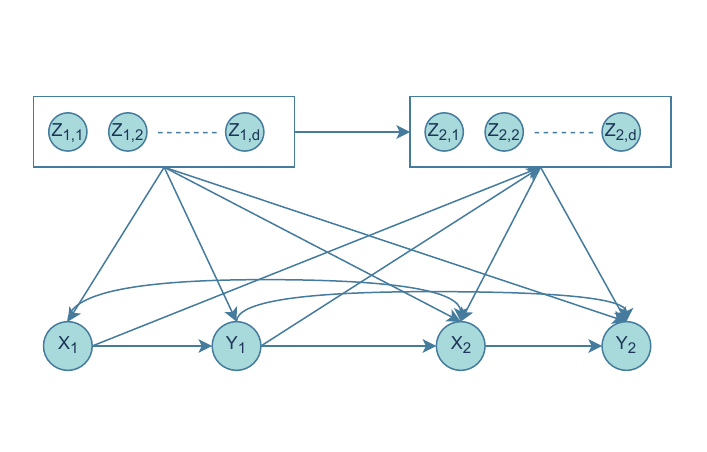}
    \caption{SCM of a Sample mSBD scenario}
    \label{msbd_SCM}
\end{figure}
\subsubsection{\textbf{Multi-outcome Sequential Back-door Criterion}}
The multi-outcome sequential back-door (mSBD) criterion extends sequential adjustment to ordered treatments and outcome blocks~\cite{pearl1995probabilistic,jung2020estimating}. Let $\mathbf X=(X_1,\ldots,X_n)$ be topologically ordered. Following the outcome partition in Definition 3 of Jung et al.~\cite{jung2020estimating}, set $\mathbf Y_0=\mathbf Y\setminus De(\mathbf X)$ and
\[
\mathbf Y_i=\mathbf Y\cap\bigl(De(X_i)\setminus De(\mathbf X^{\ge i+1})\bigr)
\]
for $i=1,\ldots,n$, where $\mathbf X^{\ge i}=(X_i,\ldots,X_n)$ and $\mathbf X^{\ge n+1}=\varnothing$. Write $\mathbf X^{(i)}=(X_1,\ldots,X_i)$, $\mathbf Z^{(i)}=(\mathbf Z_1,\ldots,\mathbf Z_i)$, and $\mathbf Y^{(i)}=(\mathbf Y_0,\ldots,\mathbf Y_i)$. The histories $\mathbf X^{(0)}$, $\mathbf Z^{(0)}$, and $\mathbf Y^{(-1)}$ are empty.

The covariate sequence must satisfy mSBD admissibility: each $\mathbf Z_i$ is a nondescendant of current and subsequent treatments, and the required graphical separation of each treatment from subsequent outcomes holds given the past outcomes, past treatments, and available covariates in the modified graph specified by that definition. Merely assigning time indices is insufficient. Under this criterion,
\begin{multline}
P_{\mathbf x}(\mathbf y)=\sum_{\mathbf z}
\prod_{k=0}^{n}p(\mathbf y_k\mid\mathbf x^{(k)},\mathbf z^{(k)},\mathbf y^{(k-1)})\\
{}\times\prod_{j=1}^{n}p(\mathbf z_j\mid\mathbf x^{(j-1)},\mathbf z^{(j-1)},\mathbf y^{(j-1)}).
\end{multline}
The corresponding weighting identity is
\begin{equation}
\label{msbd_eq}
\begin{aligned}
\mu_h(\mathbf x)&=\beta[h(\mathbf Y)\mid\mathbf x;\omega_{\mathrm{mSBD}}],\\
\omega_{\mathrm{mSBD}}(\mathbf x,\mathbf y,\mathbf z)
&=\frac{p(\mathbf x)}{\prod_{k=1}^{n}p(x_k\mid\mathbf x^{(k-1)},\mathbf y^{(k-1)},\mathbf z^{(k)})}.
\end{aligned}
\end{equation}
In Figure~\ref{msbd_SCM}, the target used in our experiments is the final-outcome mean, obtained by taking $h(\mathbf Y)=Y_2$. The two-treatment weight specializes to
\begin{equation}
\label{msbd_weight}
\omega=\frac{p(x_1,x_2)}{p(x_1\mid\mathbf z_1)p(x_2\mid x_1,y_1,\mathbf z_1,\mathbf z_2)}.
\end{equation}

\begin{figure}[t]\centering
    \includegraphics[width=\linewidth]{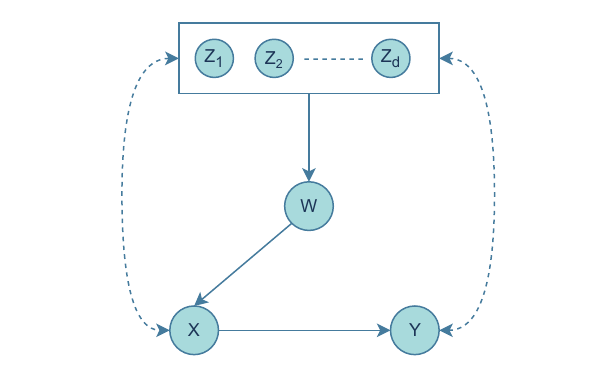}
    \caption{SCM of a Sample Surrogate scenario}
    \label{surr_SCM}
\end{figure}
\subsubsection{\textbf{Surrogate Criterion}}
The surrogate setting in Figure~\ref{surr_SCM} uses an auxiliary variable $W$ to identify the intervention effect of $X$ on $Y$. It does not replace the target treatment $X$ by $W$. The covariates $\mathbf Z$ are used to construct weights for $W$; they are not an ordinary BD adjustment set for $X\to Y$ in this graph. In particular, conditioning on the collider in $X\leftrightarrow\mathbf Z\leftrightarrow Y$ can open that path.

Assume that $(W,\mathbf Z)$ satisfies the surrogate criterion relative to $(X,Y)$ in Definition 4 of Jung et al.~\cite{jung2020estimating}, including its graphical separation conditions and mSBD admissibility relative to $(W,(X,Y))$. For the simple graph shown, Theorem 3 gives
\begin{equation}
\label{surr_eq}
\mu_h(x)=\beta[h(Y)\mid(x,w);\omega],\qquad
\omega(v)=\frac{p(w)}{p(w\mid\mathbf z)}.
\end{equation}
Conditioning on $(x,w)$ means $(X,W)=(x,w)$. Under these conditions the population expression is invariant to the admissible value of $w$. This invariance need not hold exactly for a fitted finite-sample regression.

\begin{figure}[t]\centering
    \includegraphics[width=\linewidth]{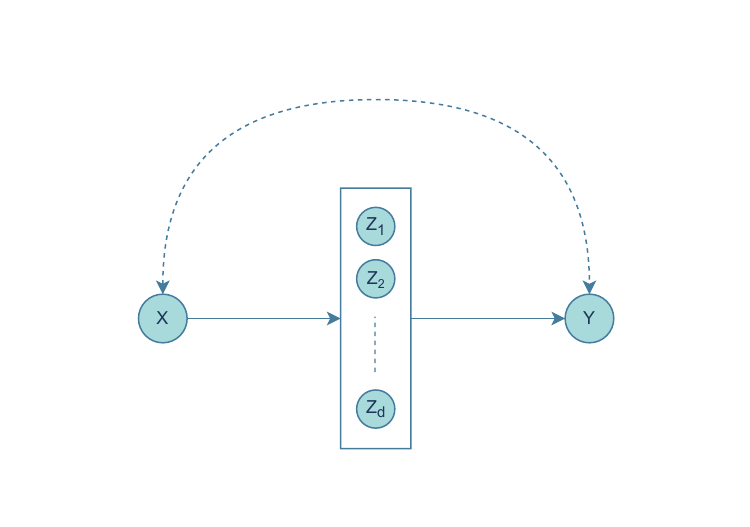}
    \caption{SCM of a Sample Front-door scenario}
    \label{fd_SCM}
\end{figure}
\subsubsection{\textbf{Front-door Criterion}}
In Figure~\ref{fd_SCM}, $X$ and $Y$ share an unobserved common cause, and $\mathbf Z$ is a mediator set. The classical front-door criterion requires that $\mathbf Z$ intercept every directed path from $X$ to $Y$, that no unblocked BD path connect $X$ to $\mathbf Z$, and that $X$ block all BD paths from $\mathbf Z$ to $Y$~\cite{pearl-2009}. Measuring a mediator without these conditions is not sufficient.

The empty set is BD admissible for the effect of $X$ on $\mathbf Z$, so the outer operator has unit weight:
\begin{equation}
\label{fd_eq1}
\beta_1[h](x)=\beta[h(\mathbf Z)\mid x;1]
=\mathbb E[h(\mathbf Z)\mid X=x].
\end{equation}
For the effect of $\mathbf Z$ on $Y$, $X$ is BD admissible. Define
\begin{equation}
\label{fd_eq2}
\begin{aligned}
g(\mathbf z)&=\beta[Y\mid\mathbf z;\omega_2],\qquad
\omega_2(v)=\frac{p(\mathbf z)}{p(\mathbf z\mid x)},\\
g(\mathbf z)&=\sum_{x'}\mathbb E[Y\mid X=x',\mathbf Z=\mathbf z]p(x').
\end{aligned}
\end{equation}
The composition yields the total interventional mean,
\[
\mu(x)=\beta_1[g](x)=\mathbb E[g(\mathbf Z)\mid X=x].
\]
It is not a direct effect holding the mediator fixed. These are the CWO front-door identities~\cite{jung2020estimating}.

% \begin{figure}[t]\centering
%     \includegraphics[width=7cm,height=4.5cm,keepaspectratio]{media.drawio.pdf}
%     \caption{SCM of a Sample Causal Mediators}
%     \label{med_SCM}
% \end{figure}

% \subsubsection{Causal Mediators}
% In causal models, there can be multiple paths from treatment \textbf{X} to outcome \textbf{Y} through different covariates  (\textbf{Z}, \textbf{Q\textsubscript{1}}, \textbf{Q\textsubscript{2}}). The importance of relative strength of causal effects through different pathways (direct or indirect) have led to the development of causal mediation analysis. We can estimate causal effects \textit{P}\textsubscript{x}(y) using the composition of two weighting operators $\beta$\textsubscript{1} and $\beta$\textsubscript{2} whenever ( \textbf{Z}, \textbf{Q\textsubscript{1}}, \textbf{Q\textsubscript{2}} ) satisfies mSBD composition criterion\cite{jung2020estimating} relative to (\textbf{X,Y}). 
% \begin{definition}
% if ( \textbf{Z}, \textbf{Q\textsubscript{1}}, \textbf{Q\textsubscript{2}} ) is mSBD admissible to (\textbf{X},\textbf{Y}) then,
% \begin{equation}
%     E[P\textsubscript{\textbf{x}}[h(\textbf{Y})]] = (\beta\textsubscript{1} o\beta\textsubscript{2})(x) 
% \end{equation}
% where, \begin{math} \beta\textsubscript{1}(x) \equiv \beta[h(\textbf{Z})|x;\mathcal{W}\textsubscript{mSBD}(x,z,q\textsubscript{1})]\end{math} \\and \begin{math}  \beta\textsubscript{2}(x) \equiv \beta[\textbf{Y}|z;\mathcal{W}\textsubscript{mSBD}(z,y,q\textsubscript{2})]\end{math}
% \end{definition}

%%%%%%%%%%%%%%%%%%%%%%%%%%%%%%%%%%%%%%%%%%%%%%%%%%%%%%%%%%%%%%%%%%%%%%%%

\section{The NN-CWO Algorithm}
\label{sec:methods}
NN-CWO approximates the regression functions in the preceding CWO identities. For input rows $A_i$, responses $B_i$, and estimated weights $\widehat\omega_i$, the weighted squared-error objective is
\[
\sum_{i=1}^{N}\widehat\omega_i\{B_i-f_\theta(A_i)\}^2.
\]
With true weights and an unrestricted function class, the population minimizer is the weighted conditional mean in Equation~\ref{weight_operator}. A finite neural network trained by Adam approximates this fit; neither the architecture nor backpropagation establishes causal identification or consistency on its own. The graph and weighting identity supply the causal interpretation.

\begin{algorithm}[t]
\SetAlgoLined
\SetKwBlock{Begin}{Function}{end function}
\Begin(NN\_CWO{(}X, y, X\_pred, w, hp{)}:)
{
$p\gets$ number of columns of X\;
\If{p $\neq$ 1}{
    \textit{Beta}$\gets$ \textbf{Sequential Neural Network}\;
    \textbf{+} Dense(``linear", hp[input\_units], p)\;
\textbf{+} Dropout( hp[dropout\_rate])\;

\For{$i\gets1$ \KwTo $hp[n\_layers]$}{
     \textbf{+} Dense(``relu", hp[units\_i])\;
    \textbf{+} Dropout(hp[dropout\_rate\_i])\;   
  }

\textbf{+} Dense(``linear", 1)\;

\textit{Beta}.compile(``adam", hp[learning\_rate], ``mean squared error")\;

\textit{Beta}.fit(X, y, sample\_weight=w)\;
}

\Else{
\textit{Beta}$\gets$\textbf{LinearRegression()} \;
\textit{Beta}.fit(X, y, sample\_weight=w)\;
}
$y\_pred\gets$ \textit{Beta}.predict(X\_pred)\;
\Return y\_pred
}
 \caption{NN-CWO Algorithm}
 \label{algo_1}
\end{algorithm}

Algorithm~\ref{algo_1} uses \textit{X} as a generic regression-input matrix, \textit{y} as the response, \textit{X\_pred} as the prediction inputs, \textit{w} as sample weights, and \textit{hp} as hyperparameters. Let $p$ denote the number of columns of this input matrix, distinct from the simulation covariate dimension $D$. The stated implementation uses a neural network for $p\ne1$ and linear regression for $p=1$. This scalar-input branch is an implementation choice, not a general claim that univariate regression functions are linear. With binary inputs and an intercept, a linear regression can represent both group means.

The pseudocode specifies an input dense layer, ReLU hidden layers with dropout, and a scalar linear output, fitted using Adam and weighted mean squared error. The hyperparameters were reported as selected by a search procedure. Backpropagation optimizes the regression fit. The function returns predictions at \textit{X\_pred}; the application-specific algorithms determine how these predictions enter the causal functional.
% VERIFY: the original prose specified a ReLU input activation, while the
% original pseudocode specified linear. The pseudocode is preserved here;
% the actual input activation must be checked in the experiment code.
% VERIFY: search spaces, selection criterion, callbacks and sample-weight API
% usage require the original implementation; no new settings are assumed.

\subsection{Front-door}
\begin{algorithm}[t]
\SetAlgoLined
\SetKwBlock{Begin}{function}{end function}
\Begin(front-door\_NNCWO{(}data, hp{)}){
$D\gets$ dimension of columns of \textbf{Z} of \textit{data}\;
$X\gets$ data[x]\;
$Y\gets$ data[y]\;
$Z\gets$ data[z\textsubscript{1} to z\textsubscript{D}]\;
$weight\textsubscript{1}\gets$ getWeight()\;
$Y\textsubscript{2}\gets$ NN\_CWO(Z, Y, Z, weight\textsubscript{1}, hp)\;
$weight\textsubscript{2}\gets$ column of ones\;
$X\_test\gets$ [0,1]\;
$y\_pred\gets$ NN\_CWO(X, Y\textsubscript{2}, X\_test, weight\textsubscript{2}, hp)\;
$\mu\textsubscript{0}\gets$ y\_pred[0]\;
$\mu\textsubscript{1}\gets$ y\_pred[1]\;

\Return $\mu$\textsubscript{0}, $\mu$\textsubscript{1}\\
}
 \caption{Intervention-specific outcome means: front-door}
 \label{fd_algo}
\end{algorithm}

Algorithm~\ref{fd_algo} expresses the two regressions implied by Equations~\ref{fd_eq1}--\ref{fd_eq2}. First, \textit{getWeight()} evaluates estimates of $p(\mathbf z)/p(\mathbf z\mid x)$ at the observations. The call \textit{NN\_CWO(Z, Y, Z, weight\textsubscript{1}, hp)} fits $Y$ on $\mathbf Z$ with these weights and returns $\widehat g(\mathbf Z_i)$, stored as \textit{Y\textsubscript{2}}. Second, a regression of these fitted values on $X$ with unit weights approximates $\mathbb E[g(\mathbf Z)\mid X=x]$. Predictions at $x=0,1$ give $\widehat\mu(0)$ and $\widehat\mu(1)$.

These outputs are intervention-specific means. Their difference is an ATE for binary treatment, but the algorithm returns the means separately, as used in the evaluation.
% VERIFY: the original printed call was (Z, X, X, ...), while the prose had
% (Z, X, Z, ...). The call above aligns the pseudocode with the existing
% identification equations. Confirm that the experiment code used (Z,Y,Z,...)
% before treating this as a verified transcription-only correction.

\subsection{Surrogate Criterion}
\begin{algorithm}[t]
\SetAlgoLined
\SetKwBlock{Begin}{function}{end function}
\Begin(surrogate\_NNCWO{(}data, hp{)}){
$D\gets$ dimension of columns of \textbf{Z} of \textit{data}\;
$X\gets$ data[x]\;
$Y\gets$ data[y]\;
$X\_train\gets$ data[x, w]\;
$weight\gets$ getWeight()\;
$X\_test\gets$ [[0,1],[1,0]]\;
$y\_pred\gets$ NN\_CWO(X\_train, Y, X\_test, weight, hp)\;
$\mu\textsubscript{0}\gets$ y\_pred[0]\;
$\mu\textsubscript{1}\gets$ y\_pred[1]\;
\Return $\mu$\textsubscript{0}, $\mu$\textsubscript{1}\\
}
\caption{Intervention-specific outcome means: surrogate}
\label{surr_algo}
\end{algorithm}

Algorithm~\ref{surr_algo} fits a weighted regression of $Y$ on $(X,W)$ using estimates of $p(w)/p(w\mid\mathbf z)$ from Equation~\ref{surr_eq}. Its justification is the surrogate identity, not ordinary BD adjustment for a joint intervention on $(X,W)$. Although the probability models involve $D$ covariates, this response regression has two input columns.

The prediction rows $(0,1)$ and $(1,0)$ are two selected assignments, not all binary combinations. Under the population surrogate identity, either admissible $w$ gives the same target for a fixed $x$. The algorithm retains these two rows and returns $\widehat\mu(0)$ and $\widehat\mu(1)$. Their finite-sample values can depend on the selected rows because the fitted regression need not satisfy exact invariance. Relevant support is required; no averaging over additional assignments is introduced.

\subsection{mSBD}
\begin{algorithm}[t]
\SetAlgoLined
\SetKwBlock{Begin}{function}{end function}
\Begin(mSBD\_NNCWO{(}data, hp{)}){
$X\gets$ data[x\textsubscript{1}, x\textsubscript{2}]\;
$Y\gets$ data[y\textsubscript{2}]\;
$weight\gets$ getWeight()\;
$X\_test\gets$ [[0,0], [0,1], [1,0], [1,1]]\;
$y\_pred\gets$ NN\_CWO(X, Y, X\_test, weight, hp)\;
$\mu\textsubscript{00}\gets$ y\_pred[0]\;
$\mu\textsubscript{01}\gets$ y\_pred[1]\;
$\mu\textsubscript{10}\gets$ y\_pred[2]\;
$\mu\textsubscript{11}\gets$ y\_pred[3]\;

\Return $\mu$\textsubscript{00}, $\mu$\textsubscript{01}, $\mu$\textsubscript{10}, $\mu$\textsubscript{11} \\
}
\caption{Intervention-specific final-outcome means: mSBD}
\label{msbd_algo}
\end{algorithm}

Algorithm~\ref{msbd_algo} targets
\[
\mu(x_1,x_2)=\mathbb E[Y_2\mid do(X_1=x_1,X_2=x_2)]
\]
in Figure~\ref{msbd_SCM}. The input matrix \textit{X} contains $(X_1,X_2)$, the response is $Y_2$, and \textit{getWeight()} evaluates the estimated sequential weight in Equation~\ref{msbd_weight}. The weight includes the intermediate outcome and covariate histories even though only $Y_2$ is predicted. Evaluating \textit{NN\_CWO(X, Y, X\_test, weight, hp)} at all four binary treatment assignments returns four regime-specific means, not an unspecified ATE contrast. The response regression has two input columns; $D$ denotes the dimension of each covariate block used in weight construction.

\section{Empirical Evaluations}
\label{sec:res}
We compare NN-CWO with the CWO baseline in the front-door, surrogate, and mSBD settings. The simulations follow the corresponding causal structures and evaluation approach in Jung et al.~\cite{jung2020estimating}. Experiments were performed using Google Colaboratory. Comparisons reported in that earlier work do not establish NN-CWO's performance against methods not evaluated here.
% VERIFY: exact agreement of structural equations, parameter settings and
% baseline implementation with the earlier study requires experiment records.

\subsection{Experimental Settings}
For each treatment assignment $a$, the target is the intervention-specific outcome mean $\mu(a)$. A large interventional sample of size $10^7$ supplies a Monte Carlo approximation, denoted $\widetilde\mu_{\mathrm{MC}}(a)$, used as the reference. This is an approximation to the target, rather than an exact analytic value. The observational-sample estimates from the two methods are denoted $\widehat\mu_{\mathrm{CWO}}(a)$ and $\widehat\mu_{\mathrm{NN}}(a)$.

Weights are estimated using parametric probability models. Multivariate probabilities are factored using the chain rule, with logistic regression used for binary conditional distributions. The chain rule itself does not ensure correct specification of the fitted models. The probability-model specifications and support affect both methods.
% VERIFY: record probability-model predictors, interactions, factorization
% order, and fitting settings from the original code; do not infer them here.

Let $\mathcal A$ be the set of evaluated treatment assignments and $K=|\mathcal A|$. For repetition $r$ and method $R$, the average absolute error is
\begin{equation}
\label{aae_metric}
\operatorname{AAE}^{(r)}_R
=\frac{1}{K}\sum_{a\in\mathcal A}
\left|\widehat\mu_R^{(r)}(a)-\widetilde\mu_{\mathrm{MC}}(a)\right|.
\end{equation}
There are two evaluated assignments for the binary-treatment front-door and surrogate targets, and four for the two-binary-treatment mSBD target. We summarize the 100 reported AAE values at each sample size by
\[
\operatorname{MAAE}_R=\operatorname{median}_{r=1,\ldots,100}
\operatorname{AAE}^{(r)}_R.
\]
The metric averages errors in intervention-specific means, not errors in treatment-effect contrasts. The median summarizes typical performance and reduces sensitivity to extreme repetitions; it does not quantify uncertainty or establish statistical significance. Lower MAAE indicates lower error by this summary.
% VERIFY: clarify whether the 100 repetitions vary datasets, SCMs, network
% initializations, or combinations. If the Monte Carlo reference is regenerated
% per repetition, add an r index to that reference as well.
% VERIFY: the original abstract reports up to 55% improvement. Confirm the
% metric and particular comparison before describing it as relative MAAE
% reduction, 100*(MAAE_CWO-MAAE_NN)/MAAE_CWO.

\subsection{Results}
We make estimations for the different scenarios over a range of dimensions. For efficient estimation, we perform all the experiments over a range of samples \{500, 1000, 1500, ..., 10000\}. Here, dashed lines in a graph represents the results for CWO algorithm and solid lines represents the results for NN-CWO algorithm.

	\begin{figure}[t]
		\begin{subfigure}[h]{\linewidth}
			\centering
			\includegraphics[width=\linewidth]{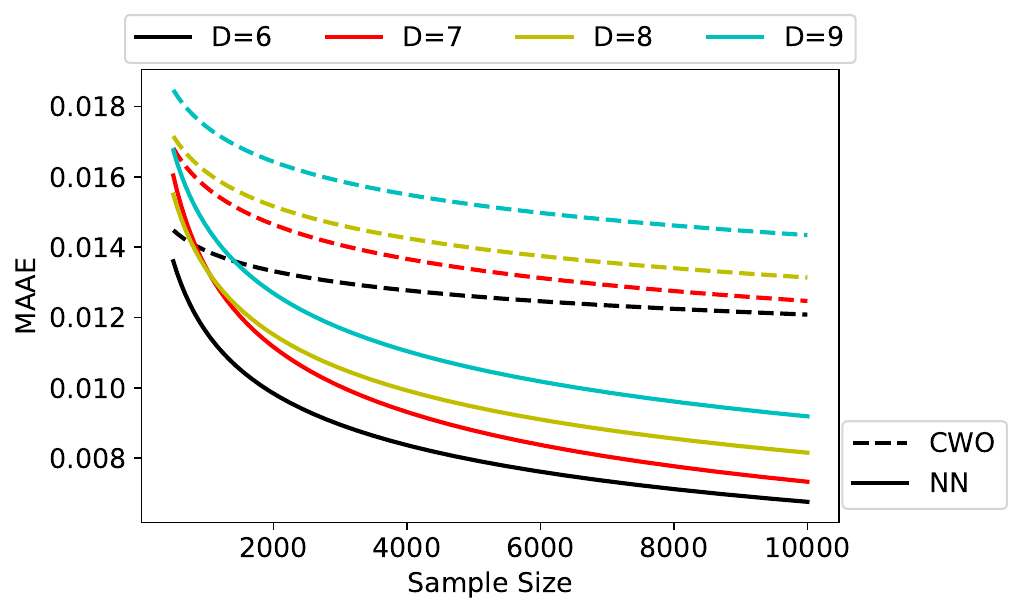}
			\hspace{-4mm}
			\caption{low dimensions}
			\label{fd_graph_low}
			\Description{}
		\end{subfigure}
		\begin{subfigure}[h]{\linewidth}
			\centering
			\includegraphics[width=\linewidth]{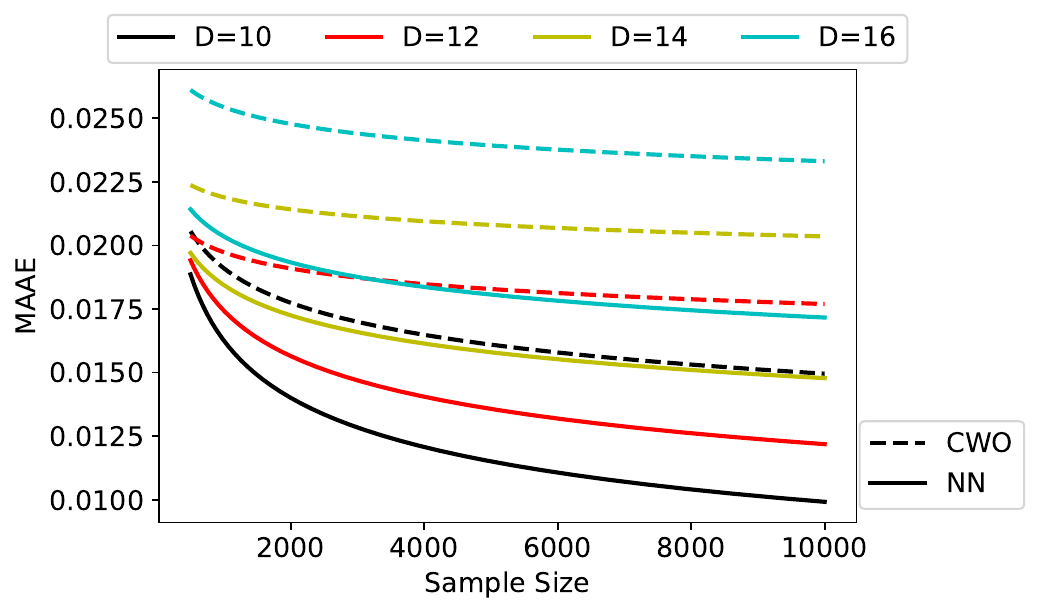}
			\vspace{-2mm}
			\caption{high dimensions}
			\label{fd_graph_high}
			\Description{}
		\end{subfigure}
		
		\caption{MAAE plots for Front-door Criterion}
        \label{f_door}
\end{figure}

\textbf{Front-door criterion.} Treatment $X$ is binary, $Y$ is continuous in $[0,1]$, and $\mathbf Z=(Z_1,\ldots,Z_D)$. Figures~\ref{fd_graph_low} and~\ref{fd_graph_high} show the lower- and higher-dimensional settings. NN-CWO has lower plotted MAAE than CWO for the evaluated sample sizes and dimensions. Errors generally decline as sample size grows, with differences between dimensions and methods. These finite-sample comparisons do not establish asymptotic consistency or show that dimension necessarily induces nonlinearity.

	\begin{figure}
		\begin{subfigure}[h]{\linewidth}
			\centering
			\includegraphics[width=\linewidth]{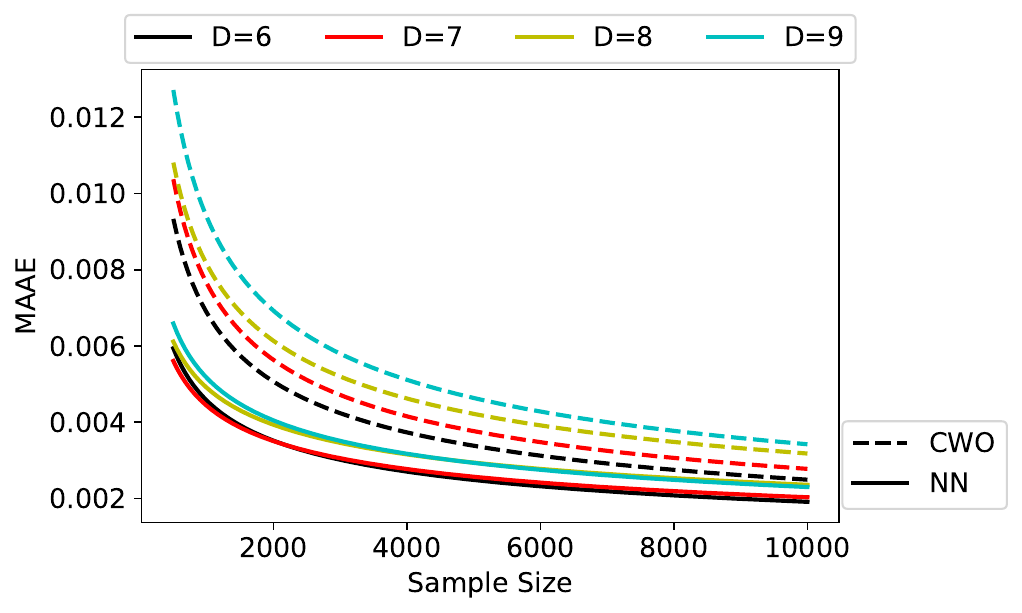}
			\vspace{-2mm}
			\caption{low dimensions}
			\label{surr_graph_low}
			\Description{}
		\end{subfigure}
		\begin{subfigure}[h]{\linewidth}
			\centering
			\includegraphics[width=\linewidth]{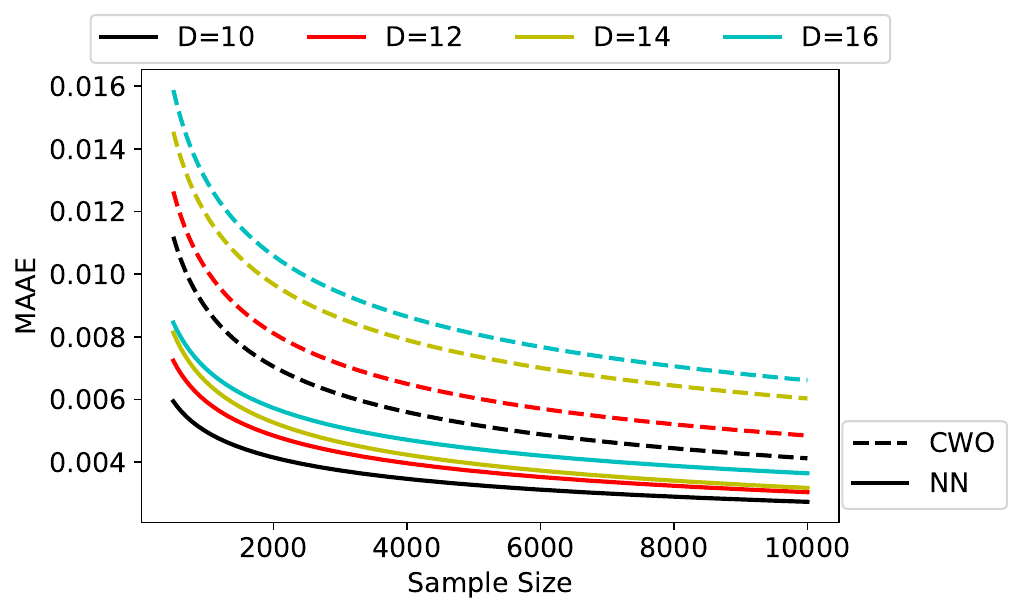}
			\vspace{-2mm}
			\caption{high dimensions}
			\label{surr_graph_high}
			\Description{}
		\end{subfigure}
		\caption{MAAE plots for Surrogate Criterion}
\end{figure}
\textbf{Surrogate criterion.} We estimate $\mu(x)$ for Figure~\ref{surr_SCM} using Algorithm~\ref{surr_algo}. Variables $X$, $W$, and each component of $\mathbf Z$ are binary; $Y$ is continuous in $[0,1]$, and $D$ is the dimension of $\mathbf Z$. Figures~\ref{surr_graph_low} and~\ref{surr_graph_high} show lower plotted MAAE for NN-CWO in the evaluated settings. Both methods' errors generally decrease with sample size. A lower error level does not imply a larger absolute decrease: several CWO curves start higher and decline more steeply while remaining above NN-CWO.

\begin{figure}[t]
			\centering
			\includegraphics[width=\linewidth]{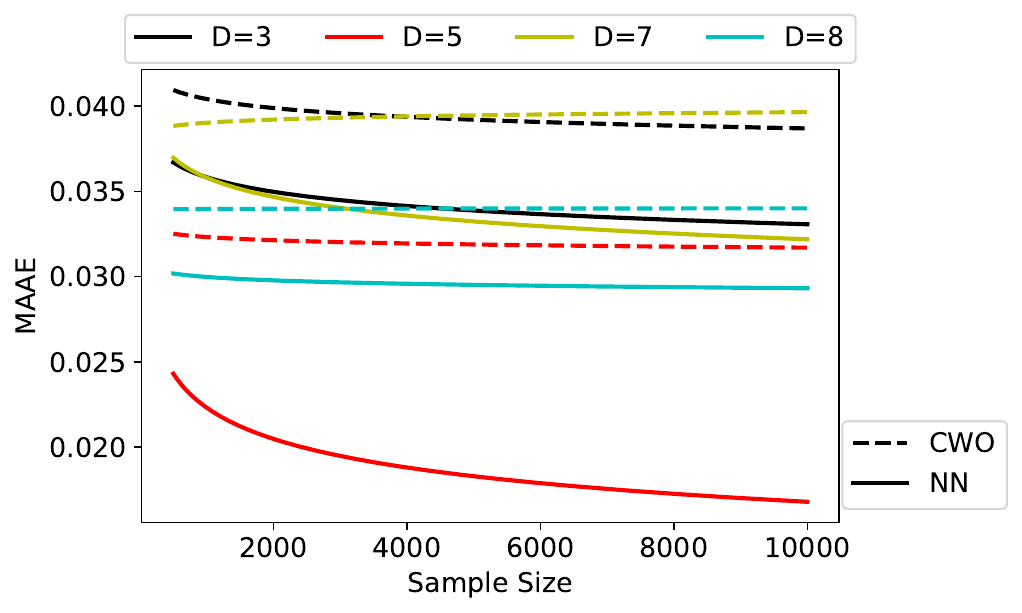}
			
			\caption{MAAE plot for mSBD criterion}
			\label{msbd_graph}
			
		\end{figure}
\textbf{mSBD criterion.} We estimate $\mu(x_1,x_2)$ for the graph in Figure~\ref{msbd_SCM} using Algorithm~\ref{msbd_algo}. The treatment variables $X_1,X_2$, the intermediate outcome $Y_1$, and the components of the $D$-dimensional covariate blocks $\mathbf Z_1,\mathbf Z_2$ are binary; $Y_2$ is continuous. Figure~\ref{msbd_graph} shows lower plotted MAAE for NN-CWO across the evaluated settings. The curves have different trends and do not uniformly decrease; the dashed $D=7$ curve increases slightly. The plots do not establish vanishing error or isolate the mechanism responsible for the performance difference.

The comparisons concern these simulated settings and the CWO baseline. They do not demonstrate robustness to misspecified causal graphs, unmeasured confounding beyond that permitted by the identifying conditions, or misspecified weight models. No uncertainty assessment for the differences is reported.

\balance
\section{Conclusions and Future Works}
\label{sec:conc}
We studied NN-CWO, an empirical implementation of CWO using neural-network regression for its identified statistical functionals. The reported front-door, surrogate, and mSBD simulations show lower MAAE than the CWO baseline in the evaluated settings. These results concern estimation under assumed causal structures, rather than discovery of causal relationships or a general guarantee of superiority. Future work includes assessing model misspecification, evaluating mSBD composition, and investigating recurrent architectures for richer longitudinal settings beyond the sequential-treatment example considered here.

%%%%%%%%%%%%%%%%%%%%%%%%%%%%%%%%%%%%%%%%%%%%%%%%%%%%%%%%%%%%%%%%%%%%%%%%

%%% The next two lines define, first, the bibliography style to be 
%%% applied, and, second, the bibliography file to be used.
\bibliographystyle{ACM-Reference-Format} 
\bibliography{sample}

%%%%%%%%%%%%%%%%%%%%%%%%%%%%%%%%%%%%%%%%%%%%%%%%%%%%%%%%%%%%%%%%%%%%%%%%

\end{document}